\RequirePackage[2020-02-02]{latexrelease}

\documentclass{article}

\usepackage{microtype}
\usepackage{graphicx}
\usepackage{booktabs} 

\usepackage{hyperref}

\usepackage[accepted]{preprint_custom}

\usepackage{amsmath}
\usepackage{amssymb}
\usepackage{mathtools}
\usepackage{amsthm}

\usepackage[capitalize,noabbrev]{cleveref}

\theoremstyle{plain}

\theoremstyle{definition}

\theoremstyle{remark}

\usepackage[textsize=tiny]{todonotes}

\graphicspath{{figures/}}

\preprinttitlerunning{UMM: Unsupervised Mean-Difference Maximization}

\usepackage[utf8]{inputenc} \usepackage[T1]{fontenc}    \usepackage{hyperref}       \usepackage{url}            \usepackage{booktabs}       \usepackage{nicefrac}       \usepackage{microtype}      \usepackage{marginnote}

\expandafter\let\csname equation*\endcsname\relax

\expandafter\let\csname endequation*\endcsname\relax

\usepackage{mathtools}
\usepackage{algorithm}
\usepackage{algpseudocode}
\usepackage{amssymb}
\usepackage{amsfonts}       \usepackage{blkarray}
\usepackage{siunitx}
\usepackage{subcaption}

\usepackage{xstring}
\usepackage{catchfile}

\usepackage[printwatermark]{xwatermark}
\usepackage{xcolor}
\usepackage{graphicx}
\graphicspath{{figures/}}

\usepackage{multirow}

\usepackage[textsize=tiny]{todonotes}

\usepackage{url}

\reversemarginpar

\renewcommand{\vec}[1]{\boldsymbol{\mathbf{#1}}}
\DeclareMathOperator{\mean}{mean}
\DeclareMathOperator*{\argmax}{argmax}

\DeclareMathOperator*{\cov}{cov}

\newcommand{\trans}{^\mathsf{T}}

\DeclareMathSymbol{\shortminus}{\mathbin}{AMSa}{"39}
 
\begin{document}

\twocolumn[
\preprinttitle{UMM: Unsupervised Mean-Difference Maximization}

\preprintsetsymbol{equal}{*}

\begin{preprintauthorlist}
\preprintauthor{Jan Sosulski}{uni_freiburg}
\preprintauthor{Michael Tangermann}{uni_netherlands}
\end{preprintauthorlist}

\preprintaffiliation{uni_freiburg}{Department of Computer Science, University of Freiburg, Freiburg, Germany}
\preprintaffiliation{uni_netherlands}{Donders Institute for Brain, Cognition and Behaviour, Radboud University, Nijmegen, The Netherlands}

\preprintcorrespondingauthor{Michael Tangermann}{michael.tangermann@donders.ru.nl}

\preprintkeywords{Brain signal classification, linear discriminant analysis, high dimensional covariance estimation, block-Toeplitz matrix, spatiotemporal data}

\vskip 0.3in
]

\printAffiliationsAndNotice{} 

\begin{abstract}
   Many brain-computer interfaces make use of brain signals that are elicited in response to a visual, auditory or tactile stimulus, so-called event-related potentials (ERPs).
In the predominantly used visual ERP speller applications, sets of letters shown on a screen are flashed randomly, and the participant attends to the target letter they want to spell.
When this letter flashes, the resulting ERP is different compared to when any other non-target letter flashes, and by using a sequence of binary classifications of the observed ERP responses, the brain-computer interface can detect which letter was the target.
We propose a new unsupervised approach to detect the attended letter.
In each trial, for every available letter our approach makes the hypothesis that it is in fact the attended letter, and calculates the ERPs based on each of these hypotheses.
By leveraging the fact that only the true hypothesis produces the largest difference between the class means, we can detect the attended letter.
Note that this unsupervised method does not require any changes to the underlying experimental paradigm and therefore can be employed in almost any ERP-based setup.
To deal with the very noisy electroencephalogram data, we use a block-Toeplitz regularized covariance matrix to model the background activity.
We implemented the proposed novel unsupervised mean-difference maximization (UMM) method and evaluated it in offline replays of brain-computer interface visual speller datasets. For a dataset that used 16 flashes per symbol per trial, UMM correctly classifies 3651 out of 3654 letters ($99.92\,\%$) across 25 participants.
In another dataset with fewer shorter trials, 7344 out of 7383 letters ($99.47\,\%$) are classified correctly across 54 participants with two sessions each.
Even in more challenging datasets obtained from patients with amyotrophic lateral sclerosis ($77.86\,\%$) or when using auditory ERPs ($82.52\,\%$), the obtained classification rates obtained by UMM are competitive.
As an additional benefit, stable confidence measures are provided by this novel method, which can be used to monitor convergence of UMM.

 \end{abstract}

\section{Introduction}
\label{sec:introduction}

A brain--computer interface (BCI) allows its user to control applications and devices using their brain activity~\cite{wolpaw2012brain-computer}.
Popular BCIs are visual spellers~\cite{farwell1988talking,halder2015brain-controlled,lin2018novel,verbaarschot2021visual}, wheelchair control~\cite{li2013hybrid} or even BCI-assisted rehabilitation paradigms for neurological diseases~\cite{mane2020bci,musso2022aphasia}.
A common component in these applications is the detection of a particular brain signal.

A popular signal for BCI is the event-related potential (ERP)~\cite{jung1998analyzing,blankertz2011single-trial,haider2017application}.
This signal is elicited in response to a usually external stimulus and can be detected using, e.g., the electroencephalogram (EEG)~\cite{nunez2012electric}.
The ERP response looks different for stimuli the participant attends to compared to the stimuli they ignore.
This allows a binary selection, and many BCIs use a sequence of binary decisions to allow selection from multiple classes.

A common ERP-based BCI is a visual speller application.
Here, a screen usually presents symbols in a grid layout on a screen.
In fixed intervals, a subset of these symbols is highlighted, while others remain static.
When a symbol is highlighted, that the participant currently attends to, a so-called \emph{target} ERP is elicited, whereas the other, ignored symbols elicit \emph{non-target} ERPs.
After a number of highlighting sequences of different sets of symbols, the BCI can infer the attended symbol.

As brain signals are participant-specific~\cite{jayaram2016transfer} and can change from session to session, machine learning methods are employed to reliably discern individual target and non-target ERP responses.
Popular methods~\cite{lotte2018review} in ERP-based BCI include shrinkage linear discriminant analysis (sLDA)~\cite{blankertz2011single-trial}, classifiers leveraging Riemannian geometry~\cite{barachant2014plugplay} and even neural network approaches~\cite{cecotti2010convolutional}.
Generally, these approaches require a so-called calibration phase, in which the participant is instructed which stimulus to pay attention to.
Using the labeled data obtained from the calibration phase, the classification algorithm is trained to identify the characteristics of the target and non-target ERPs as well as of the confounding noise. In LDA-based classifiers, both the signal and noise information is exploited for subsequent prediction,~i.e., the productive online usage of the speller application.
However, it is often unclear how much calibration data is required. Please note that in the calibration period, the participant cannot use the BCI productively.

There exist some unsupervised approaches that make the calibration phase superfluous, enabling the participant to immediately start using the BCI.
For example, expectation maximization can be used to find the target/non-target ERP responses~\cite{kindermans2012bayesian}.
Alternatively, a slight modification of a visual speller paradigm can enable learning by label proportions~\cite{hubner2017learning} or one can even combine both~\cite{verhoeven2017improving,hubner2018unsupervised} into a mixed approach.
While there is no calibration phase in these examples, the approaches typically require a few trials worth of data (e.g., around 7 letters in~\cite{hubner2018unsupervised}) to reach a satisfactory performance.
Each of these previous approaches use unsupervised learning to obtain the target and non-target ERP responses.
These serve as the class means for an LDA classifier, which is then used to classify each individual epoch of a trial to obtain the multi-class prediction (i.e., which letter was attended) by aggregating these many binary classifications.

Instead of aggregating the outcome of the binary events (target/non-target) into a multi-class decision (letter), we propose to make use of the whole trial in ERP-based BCIs, by forming each possible selection as a hypothesis and choosing the one that maximizes the distance between the hypothesized ERP target and non-target means.
This simple, yet surprisingly effective, unsupervised mean-difference maximization (UMM) method is computationally light and does not require any modifications of the underlying BCI paradigm, making deployment in current ERP-based paradigms straightforward.

\section{Methods}
\label{sec:methods}

\subsection{Preliminaries}

We consider a binary BCI ERP classification problem, where one trial consists of $N_e$ epochs, of which $N_e^+$ are targets and $N_e^-$ are non-targets.
Additionally, $N_e^+ < N_e^-$ and $N_e^+,N_e^- > 1$, both of which are true in virtually all ERP-based BCI applications.
Furthermore for each epoch $e_k, 1 \leq k \leq N_e$ we know which symbols $s$ out of the set of available symbols $S$ were flashed during the $k$-th highlight event.

In this setting, without label information---i.e., which symbol was focused by the participant---the task is to find the assignment $A^+$ such that all epochs $\{e_{k}\,|\,k \in A^+\}$ correspond to target events (i.e., the attended symbol was flashed) and all other epochs $\{e_{k}\,|\,k \in {A^-}\}$ with $A^- \coloneqq \{k\,|\,k \notin A^+\}$ are non-targets.

Without incorporating experimental constraints, enumerating assignments quickly would become prohibitive in practice, as the number of possible assignments is $\binom{N_e}{N_e^+}$,~e.g., for a common row-column speller with a 1:5 target/non-target ratio and 60 highlights per letter, this is $\binom{60}{10}=7.54\cdot10^{10}$.
However, we know that in all feasible assignments $A^+$ one letter has to be common among all assigned epochs, therefore the number of possible assignments reduces to the number $|S|$ of available symbols.

\subsection{Unsupervised Mean-difference Maximization (UMM)}
\definecolor{flatblue}{rgb}{0.0353,0.3176,0.6902}
\definecolor{flatred}{rgb}{0.8823,0.4392,0.3333}

\begin{algorithm}
\caption{Pseudocode for the basic UMM method. Variants of blue lines are described in~\Cref{sec:covariance,sec:mean}.}
\label{alg:pseudo_umm}
\begin{algorithmic}[1]
\Require available symbols $S$, epochs of $i$-th trial $E^{(i)}$
\For{every trial $i$}
\State \textcolor{flatblue}{ $\Sigma^{\shortminus 1} \gets \cov(E^{(i)})^{\shortminus 1}$} \Comment{no class labels needed}
   \State $d^* \gets \shortminus \infty$
   \For{$s$ in $S$}
   \State \textcolor{flatblue}{ $\Delta\vec{\mu}_s \gets \mean({E^{(i)}_{A^{s^+}}}) - \mean({E^{(i)}_{A^{s^-}}})$}
      \State $d \gets \Delta\vec{\mu}_s\Sigma^{\shortminus 1}\Delta\vec{\mu}_s\trans$
      \If{$d > d^*$}
         \State $d^* \gets d$
         \State $s^* \gets s$
      \EndIf
   \EndFor \Comment $s^*$ decoded symbol for trial $i$
\EndFor
\end{algorithmic}
\end{algorithm}

Let $E_{A} \coloneqq \{e_k \,|\, k \in A\}$ denote the set of epochs of a trial that are in assignment $A$, $\mean(\cdot)$ calculates the mean of its arguments and $\cov(\cdot)$ the covariance matrix of its arguments.
Thus, $\mean(E_{A^+})$ corresponds to the target ERP and $\mean(E_{A^-})$ corresponds to the non-target ERP response.
In an unsupervised setting we do not know which letter was attended, and therefore, which assignment is correct.
However, for every available symbol $s \in S$, we can construct the hypothesis that it is the attended symbol, and obtain a corresponding target assignments,~i.e., $A^{s^+}$ which contains all epochs where $s$ was highlighted (analogously, $A^{s^-}$ contains all epochs where $s$ was not highlighted).

According to the hypothesis that $s$ is the attended letter, calculate the vector $\Delta\vec{\mu}_s$ between the obtained hypothetical class means
\begin{align}
   \Delta\vec{\mu}_s = \mean(E_{A^{s^+}}) - \mean(E_{A^{s^-}}),
\end{align}
and the corresponding squared distance
\begin{align}
   d^2(s) = (\Delta\vec{\mu}_s)(\Delta\vec{\mu}_s)\trans
\end{align}
between the class means.
In case $s$ is the true attended symbol, $A^{s^+} = A^+$ contains only target epochs.
If $s$ is not the true attended symbol, at least one of $A^{s^+}$ and $A^{s^-}$ (or both) contain mixtures of non-target and target epochs.
As a result, averaging the epochs of different classes in each assignment will mix the corresponding ERPs, which in turn reduces the distance between the hypothetical class means.
Therefore, the squared distance $d^2(s)$ between the assumed class means will be maximal if our assumed symbol is the true symbol that was attended.

However, in a high-dimensional noisy setting, with correlated data and few samples to estimate the class means from, the squared distance is not reliable.
As a remedy, we propose to use the inverted global covariance matrix $\Sigma^{\shortminus 1}$, to remove the influence of correlated dimensions and dampen dimensions that have a high variance in general.
The resulting covariance-corrected distance metric
\begin{align}
   \label{eq:objective}
   d^\Sigma(s) = (\Delta\vec{\mu}_s)\Sigma^{\shortminus 1}(\Delta\vec{\mu}_s)\trans
\end{align}
is also known as the squared Mahalanobis distance.
Note that both distances are equivalent if the covariance is the identity matrix, e.g., as would be the case for whitened data.
The symbol $s^*$ that was actually attended can now be obtained by
\begin{equation}
\label{eq:umm_optimization_problem}
s^* = \argmax_{s} d^{\Sigma}(s).
\end{equation}

Note that the traditional Mahalanobis distance would require the within-class covariance matrices, i.e., label information would be necessary.
However, as shown by \citet{hubner2020from}, in some cases, the global covariance---i.e., pooling data of both classes and ignoring class-specific means---can be used instead.
Particularly, when the matrix is multiplied with a vector that points in the direction of the difference between the (unknown) class means.
In this case, the matrix-vector product is merely scaled by some factor when using the global instead of the within-class covariance.
As the same $\Sigma^{\shortminus 1}$ is used for all symbols, every distance is scaled equally,~i.e., this does not affect~\Cref{eq:umm_optimization_problem}.
Note that regardless of the hypothesis on which symbol is attended, the expected direction of the hypothesized class mean difference vector $\Delta\vec{\mu}_s$ points toward the same direction as the true class mean difference vector.
While noise may change the direction of $\Delta\vec{\mu}$, we assume that the overall reduction of the distance between class means caused by wrong assignments dominates.

\begin{figure*}[t]
   \centering
   \includegraphics[width=\textwidth]{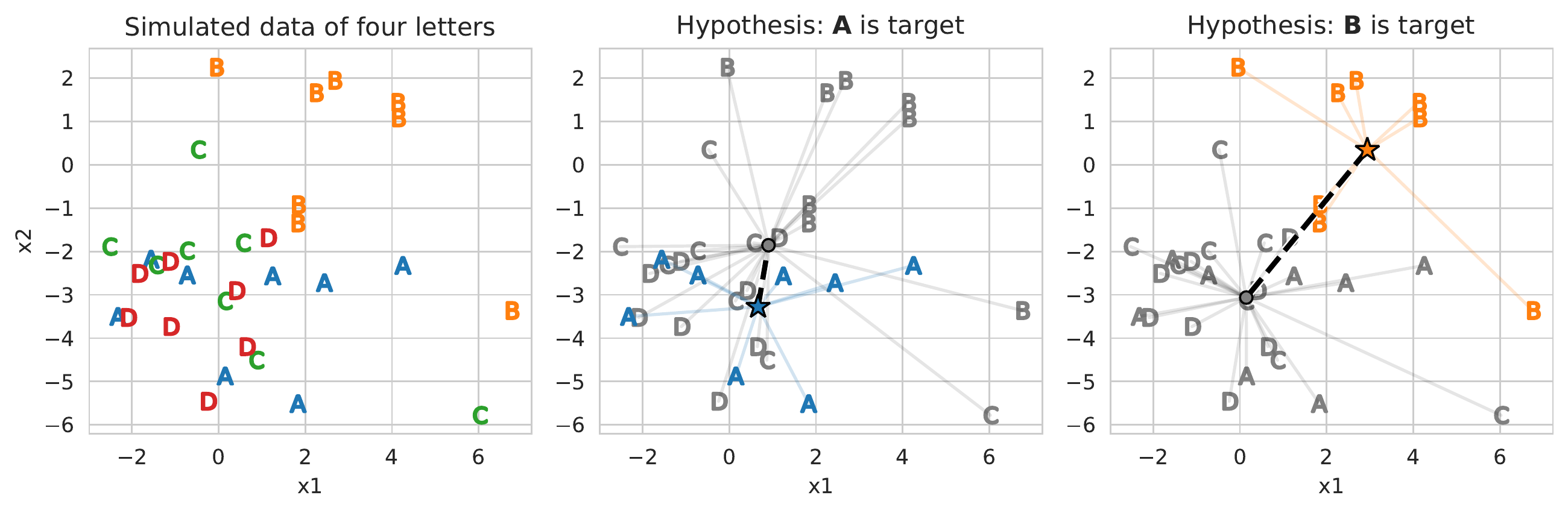}
   \caption{UMM method exemplified for two-dimensional toy data representing a four letter speller paradigm. The same multivariate Gaussian noise was assumed for all letters. The true target letters `B' (orange) were drawn from a different mean than the true non-targets `A', `C' and `D' (blue, green, red). Star markers indicate means of assumed target letters, circles of correspondingly assumed non-target letters.
\textbf{Left}: Input data (no hypothesis). \textbf{Center}:  Example of the wrong hypothesis, under which letter `A' (blue) is assumed target. Letters `B', `C', `D' are pooled to form the gray non-targets under this hypothesis. The  black dashed line indicates the difference between the hypothesized class means. \textbf{Right}: Analogously for the orange `B' being the hypothesized target class, which results in a larger distance between the hypothesized class means. Note that `C' and `D' as target hypotheses are not shown.}
   \label{fig:example_umm}
\end{figure*}

To illustrate the novel approach \Cref{fig:example_umm} shows a toy example of the unsupervised mean-difference maximization (UMM) for simulated data of a four-letter spelling problem.
Each letter is hypothesized as the target (`C' and `D' as the target are not shown), but only the true target `B' produces the largest vector between the class means.

The basic UMM method as pseudo-code is described in~\Cref{alg:pseudo_umm}.

\subsubsection{Confidence}

While our method assumes the attended symbol is the one that maximizes~\cref{eq:objective}, we can make use of all distance values generated by the other hypotheses to define a notion of UMM's confidence.

After UMM determined $s^*$ to be the attended symbol, let $S^{-}$ describe the set of all other symbols,~i.e., ${S^{-}} \coloneqq \{s\,|\,s \neq s^*, s \in S\}$ with corresponding distances $D^\Sigma_{S^{-}} \coloneqq \{d^\Sigma(s)\,|\, s \in S^{-} \}$.

Now $D^\Sigma_{S^{-}}$ allows to calculate the standard deviation $\sigma_{S^{-}}$ of the class mean distances of the presumably not attended symbols.
It can be used to standardize a comparison of the distance produced by the symbol assignment $s^*$ with the runner-up assignment $s^r$:
\begin{equation}
   \label{eq:calc_conf}
   c = \dfrac{d^\Sigma(s^*) - d^\Sigma(s^r)}{\sigma_{S^{-}}},
\end{equation}
where the runner-up is determined by $ s^r = \argmax_{s} d^\Sigma(s)$ where $s \neq s^*$.
   
The obtained confidence value $c$ for this choice of $s^*$ is always positive or zero.
Intuitively, if differences in distances is only caused by Gaussian noise then the distance between the winner and the runner-up should remain small. A large confidence on the other hand is unlikely to be caused merely by noise, instead UMM's decision for $s^*$ is more likely to be correct.

\subsubsection{Learning Across Trials}
\label{sec:mean}

So far, UMM is applied instantaneously---i.e., using only the epochs of the trial at hand---and did not incorporate any information of previous trials.
Making use of data of previous trials is straightforward for the covariance estimation.
Instead of using only the epochs of the current $i$-th trial $\Sigma^1 \coloneqq \cov(E^{(i)})$ (cf.~\Cref{alg:pseudo_umm}, line 2), we pool the data of the current trial and all previous trials, i.e., $\Sigma^{all} \coloneqq \cov(E^{(1)} \cup \ldots \cup E^{(i-1)} \cup E^{(i)})$.

Correspondingly, class mean estimates obtained from the current trial only might be improved by replacing them by a more robust estimate that makes use of previous trials (cf.~\Cref{alg:pseudo_umm}, line 5).
It makes use of the weighted average between the class mean estimates of the previous $N_t$ trials and the estimate obtained from the current trial,~i.e.,
\begin{equation}
   \label{eq:optimistic}
   \vec{\mu}^O_{s^+} = \dfrac{\vec{\mu}_+^{\text{prev}} \cdot N_t + \mean\left(E^{(i)}_{A^{s^+}}\right)}{N_t + 1},
\end{equation}
with $\vec{\mu}^{\text{prev}}_+$ being the target mean of previous trials. The non-target means can be calculated analogously.
Note that we are considering an unsupervised setting, and therefore we do not know the true class means/labels of previous trials.

We propose two different options to deal with this: The first option given by~\Cref{eq:optimistic} is an optimistic one, as it simply assumes that all of UMM's predictions in previous trials have been correct.
Alternatively, the previously derived confidence measure can be used to weigh the means according to their confidence,~i.e.,
\begin{equation}
   \label{eq:confidence}
   \vec{\mu}^C_{s^+} = \dfrac{\left[\sum\limits_{l=1}^{N_t}(\hat{c}^{(l)} \cdot \vec{\mu}^{(l)}_+) + c^{(i)} \cdot \mean\left(E^{(i)}_{A^{s^+}}\right)\right]}{\sum\limits_{l=1}^{N_t}(\hat{c}^{(l)})+c^{(i)}} ,
\end{equation}
where $c^{(l)}$ is the confidence and $\vec{\mu}^{(l)}$ the mean obtained of the (already recorded) $l$-th trial, and $\hat{c}^{(l)} = \min(c^{(l)}, 1)$.
Limiting previous confidence values to $1$ is needed, because---as we show in~\Cref{sec:conf_stim}---UMM is sensitive to the specific stimulation sequence used for a symbol. Note that $c^{(i)}$ is the confidence of UMM for the current $i$-th trial, which cannot be known before calculating $\vec{\mu}_{s^+}^{C}$, therefore this $c^{(i)}$ is derived using instantaneous, i.e., only the current trial only (cf.~\Cref{eq:calc_conf}).

Note that both approaches make use of na\"ive labeling~\cite{kuncheva2008case-study}, i.e., they use their own past classification decisions and assume them to be the true labels. As such, UMM is dependent on correct classifications, especially during the very first trials.

\subsubsection{Stationary Covariance Matrix of Background Activity}
\label{sec:covariance}

To have less epochs than feature dimensions in the first few trials is prohibitive for using the vanilla sample covariance matrix.
Thus a shrinkage regularized covariance matrix~\cite{ledoit2004well-conditioned} denoted by $\Sigma_s$ may be beneficial.
Alternatively, a recently introduced block-Toeplitz structured covariance matrix~\cite{sosulski2022introducing} denoted by $\Sigma_t$ can be used.
It assumes EEG background activity to be stationary in short epochs~\cite{cohen1977stationarity}.
The authors observed improved ERP classification performance for this regularization---especially when few data is available---due to its sample efficiency.
In addition, $\Sigma_t$ reduces the required memory by a factor of $N_t$ compared to $\Sigma_s$, and more efficient algorithms are known for mathematical operations performed on block-Toeplitz matrices. Comparisons between $\Sigma_s$ and $\Sigma_t$ in our work, however, will only focus on potential classification performance differences.

\subsubsection{BCI Datasets}
\label{sec:datasets}

We evaluated our method on five publicly available BCI datasets.
For the \textbf{Hüb17} dataset by \citet{hubner2017learning}, a copy spelling task was performed by 13 healthy participants, and 31-channel EEG was recorded using gel-based electrodes to reflect the visual ERP responses.
Each participant had to spell a 63 letter sentence three times.
To spell one letter, 68 visual highlighting events were performed, where a pseudo-random set of letters was highlighted, with 16 target and 52 non-target events per letter.

The \textbf{Hüb18} dataset by \citet{hubner2018unsupervised} had almost the same setup for 12 healthy participants, except the sentence to copy spell consisted of 35 letters only.
Data obtained afterwards under a later free-spelling condition was not evaluated.

\citet{lee2019eeg} recorded the \textbf{Lee19} dataset of 54 healthy participants with two sessions of a visual ERP paradigm each. Per session, participants performed 33 letters of standard copy spelling first (used as calibration data) followed by an online block of spelling a known 36 letter sentence.
However, the UI did not indicate the next letter on the screen during the spelling process, so participants had to remember the sentence and their current position in the sentence.
While 62-channel EEG had been recorded, we included only the same 32 channels used by the original authors in their ERP classification pipeline.
In addition to using pseudo-random sets of letters during each highlighting event, the authors overlayed symbols by a familiar face to potentially evoke an additional N400f ERP response~\cite{kaufmann2011flashing}.
In this dataset 60 epochs per letter are available, with 10 target and 50 non-target epochs.

In the \textbf{Ric13} dataset, \citet{riccio2013attention} recorded 35 letters per participant for a visual ERP protocol. The eight participants were ALS patients.
The EEG was recorded at eight channels and the spelling application had a classic row-column layout.
Each letter used 120 highlighting events with 20 targets and 100 non-targets.
Note that for the UMM method, compared to pseudo-random symbol highlights, the row-column paradigm is harder to classify, when using mean estimation methods that make use of past trials.
This can be explained in an example, consider the setting that UMM chooses a wrong symbol already in the first trial.
If the correct symbol is on a different row and column than the correct letter, the hypothesis for the target assignment now contains no actual targets, whereas the non-target assignments contains all 20 target and the remaining 80 non-targets.
In this case, the vector between the means does not point towards the target but the opposite direction.
As UMM considers distances only, it is not able to detect this wrong orientation.
Using this wrong direction in the following letter would pull future means in this wrong direction.
Since the row-column interface is still popular we chose to include this more challenging paradigm into our evaluation.

Finally, the \textbf{Sch14} dataset by \citet{schreuder2014towards} is a dataset using auditory evoked ERPs in the AMUSE paradigm.
The 21 healthy participants used an auditory BCI to spell a sentence, but in contrast to the other datasets, had to correct for spelling mistakes, i.e., if the online classifier decoded a wrong selection, participants had to select an `undo' operation to correct this.
The spelling application involved a two-step procedure to select a letter:
Per step, a participant focused on one out of six different tones which were presented from six loudspeaker directions.
First the group of letters containing the target was selected and then the actual letter to be spelled.
As a wrong selection in the first step changes the required selection for the second step (i.e., from the actual letter to choosing `undo'), we could evaluate UMM's performance per selection step with respect to the correct selection of one out of six loudspeaker directions, but not regarding the correct selection of a letter.
One selection step consisted of 90 epochs with 15 target and 75 non-target tones.
The dataset contains EEG recorded at 32 channels.
Note that similarly to the \textbf{Ric13} dataset, this is a challenging paradigm, as wrong decisions create mean-difference vectors that point into the opposite direction.

Example code how to use UMM on two visual ERP datasets is available at our repository at: \newline \url{https://github.com/jsosulski/umm_demo}.

\section{Statistical Testing}

In order to compare the different mean and covariance estimation methods, we use the paired t-test, which requires the obtained average classification performances to come from a normal distribution.
As these averages are calculated from total of 108 participants, we assume that due to the central limit theorem that the calculated sample means follow a normal distribution~\cite{lumley2002importance}.
While we could employ a non-parametric paired Wilcoxon rank sum test, this test would treat a classification rate difference of, e.g., $0.98$ to $0.99$ in the same way as $0.54$ to $0.99$ and we want to penalize cases where UMM underperforms severely.

To correct for multiple testing, we used Bonferroni correction to correct for multiple testing six times and tested against a significance level of $1\,\%$.
 
\section{Results}
\label{sec:results}

We show results for three different mean estimation methods: using only the single current trial ($\mu^1$) and then using information from previous trials. First, using an optimistic estimation ($\mu^O$) and second using a confidence-based estimation ($\mu^C$).
The covariance matrix is estimated either only on the single current trial ($\Sigma^1$) or on the current and all previous trials ($\Sigma^{all}$) using either shrinkage ($\Sigma_s$) or block-Toeplitz ($\Sigma_t$) regularization.

\subsection{Effectiveness of UMM}

\begin{figure}[t]
   \centering
   \includegraphics[width=\columnwidth]{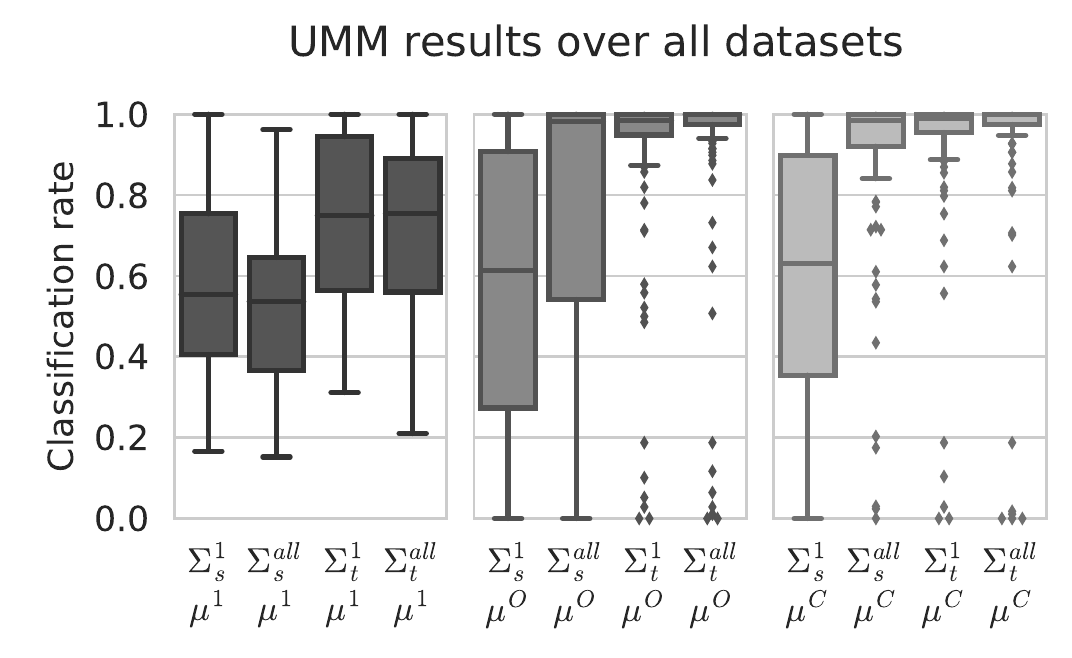}
   \caption{Boxplot of UMM classification rates for different mean and covariance estimators, pooled across all datasets. Whiskers show 1.5 interquartile ranges and diamonds indicate performance of subjects who are more than 1.5 interquartile ranges away from the first quartile.}
   \label{fig:boxplot_grand_results}
\end{figure}

Performance values of the different mean and covariance estimators in the UMM algorithm are shown in~\Cref{fig:boxplot_grand_results}.
For this plot, the average classification rate for each participant was calculated, and then all 108 participants across all datasets were pooled.
While this emphasizes the Lee19 dataset with 54 participants, we can get an overview over the performance values of the different estimators.

First off, the median classification rate of UMM for the setting with Toeplitz covariance from all past trials ($\Sigma_t^{all}$) and confidence-based mean estimation ($\mu^C$), has a median performance of $1.0$, as 58 (more than half of 108) could be classified perfectly from the very first trial on.
The difference of using Toeplitz covariance compared to shrinkage covariance is not significant when using confidence-based mean, but it is significant when using optimistic mean.
An explanation could be that Toeplitz covariance reduces the number of wrong classifications especially early on, which is important when using optimistic mean estimation and relying on early classification results.
In contrast, confidence-based mean estimation discounts the importance of wrong early classifications (if their confidence is low).
Interestingly, only shrinkage covariance estimation benefits significantly from using more than one trial of data to estimate the covariance matrix and not when using the Toeplitz covariance estimation.
For three participants of the Ric13 dataset, the classification performance is $0.0$ (the theoretical chance level is $1/36$) when using with $\mu^C$ and $\Sigma^{all}_t$ strategies.
A possible explanation could be that in these cases, UMM was initialized unfavorably and was unable to recover (cf.~\Cref{sec:datasets}).
Further evidence for this explanation is that this does not occur when using the instantaneous mean ($\mu^1$).
Using $\mu^1$ should be combined with using the instantaneous covariance estimation ($\Sigma^1$) and not all pooled data ($\Sigma^{all}$), however, this effect is only significant when using shrinkage covariance estimation.

\begin{figure*}[t]
   \centering
   \includegraphics[width=\textwidth]{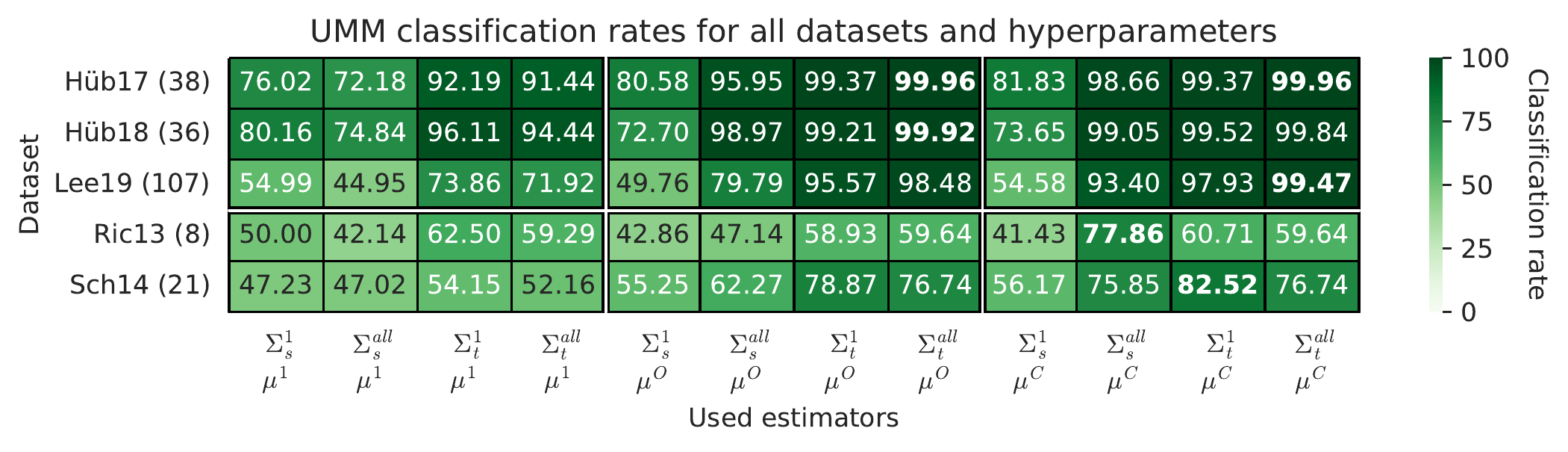}
   \caption{Heatmap with the classification rate of UMM for all datasets. Each row describes a dataset with the different covariance and mean estimators indicated in each column. Note that the top three rows correspond to the three similar visual speller datasets with pseudo-random stimulation sequences. Numbers in brackets after each dataset name indicate how many times UMM was used to spell a sentence in total, i.e., per session / block.}
   \label{fig:heatmap_all}
\end{figure*}

\Cref{fig:heatmap_all} compares UMM classification performances across the different mean and covariance estimators for each respective dataset.
As expected, the domain-specific block-Toeplitz covariance is better than using just the shrinkage covariance regularization for most datasets.
Generally, the confidence-based mean estimation performs best across all datasets.
Only the Hüb18 dataset is marginally better with the optimistic mean estimate, which corresponds to one instead of two wrong letters out of 1260 letters.
Using UMM, on all three visual speller datasets with healthy participants, classification rates above $99\,\%$ are achieved, without using any calibration data at all.
For the Ric13 dataset, a normal shrinkage covariance is better than using the Toeplitz structured matrix.
Surprisingly, in the Sch14 dataset, using a covariance matrix calculated from a single trial appears to perform better than using all the data available.
However, this occurs only due to chance as UMM is sensitive to a correct classification in the first few trials, as it learns from its own classification results when using optimistic or (less so) confidence-based class mean estimation.
Interestingly, when using only the current trial to estimate the mean ($\mu^1$), it is actually detrimental in every dataset to calculate the covariance matrix on previous trials ($\Sigma^{all}$).
A possible explanation could be, that all datasets had a fixed stimulus onset asynchrony (SOA).
As a result, in each trial the frequency that is synchronized with the SOA (e.g., 4\,Hz for SOA 250\,ms) is barely represented in the temporal covariance.

\subsection{Learning Curves}

\begin{figure}
   \centering
   \includegraphics[width=\columnwidth]{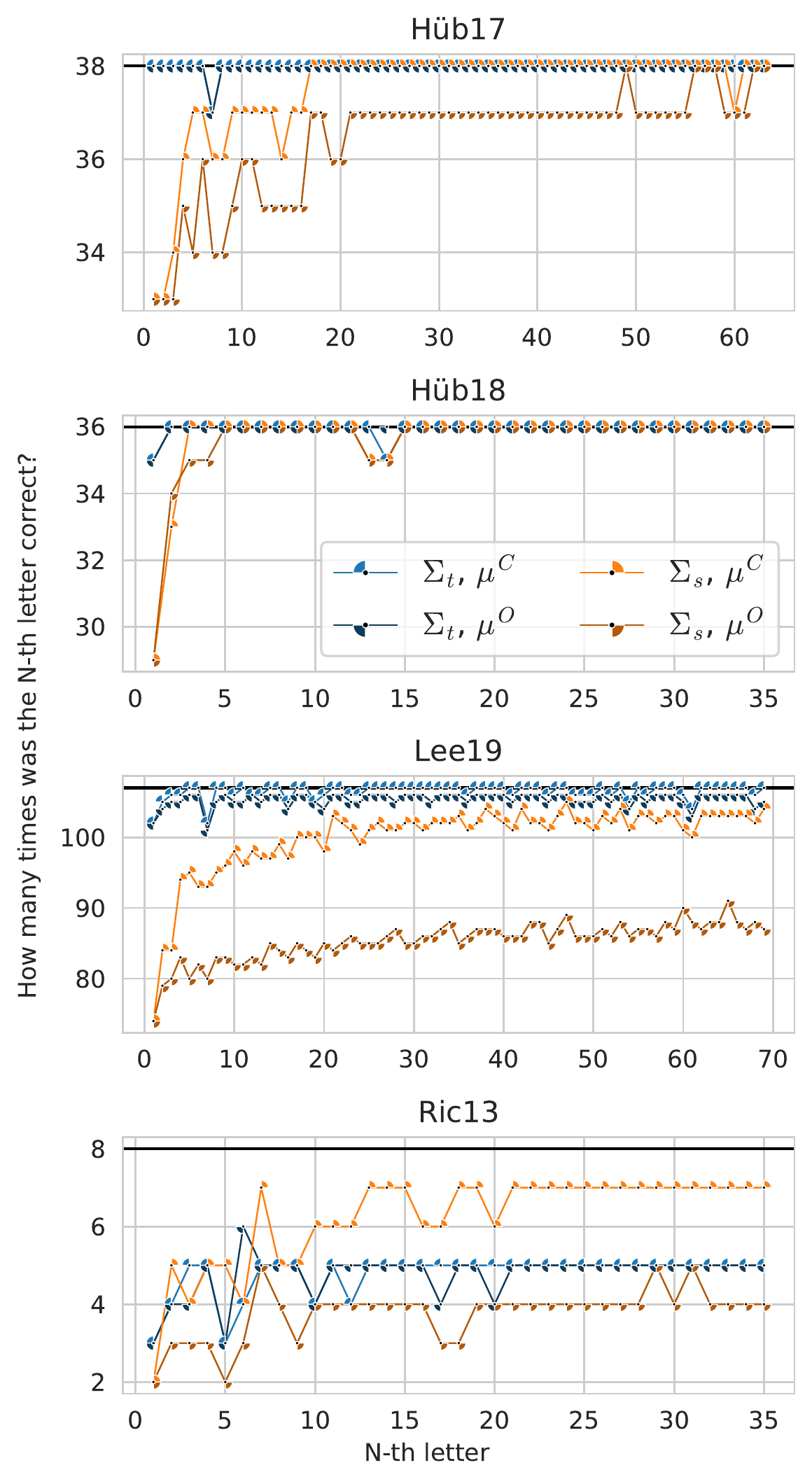}
   \caption{Learning curves for UMM using different covariance regularization and mean estimation methods for all visual speller datasets. UMM used the current and all previous trials to predict the letter. A ratio of 1 indicates that the N-th letter was correctly predicted for all sessions/participants contained in a dataset. Please note, that results for the Ric13 dataset were obtained from the recordings of 8 patients only, while the other datasets contain the results of between 36 to 107 recordings. The solid black line indicates the maximally possible performance level for each dataset.}
   \label{fig:learning_curves}
\end{figure}

The learning curves in~\Cref{fig:learning_curves} show that the Toeplitz covariance matrix (blue curves) tends to be a better choice than only the shrinkage regularized covariance for all datasets except for the Ric13 dataset.
This is especially true during the early letters, where for Hüb17 and Hüb18 there is virtually no ramp up period observed anymore with the Toeplitz regularization.
Furthermore, in all datasets, using UMM with a confidence-based mean estimation (the brighter curves within one color) outperforms the optimistic UMM approach, especially so for the Lee19 dataset when used in combination with a shrinkage estimation covariance matrix.
For the Ric13 dataset, we observed a surprisingly short ramp up period at the start, however the performance does not reach the upper ceiling as for the other datasets.
This is explained by the effect explained in the next section, where UMM can fail to perform at all if the initial trials are classified wrongly.

\subsection{Participant-wise Results for Sch14}

\begin{figure}
   \centering
   \includegraphics[width=.8\columnwidth]{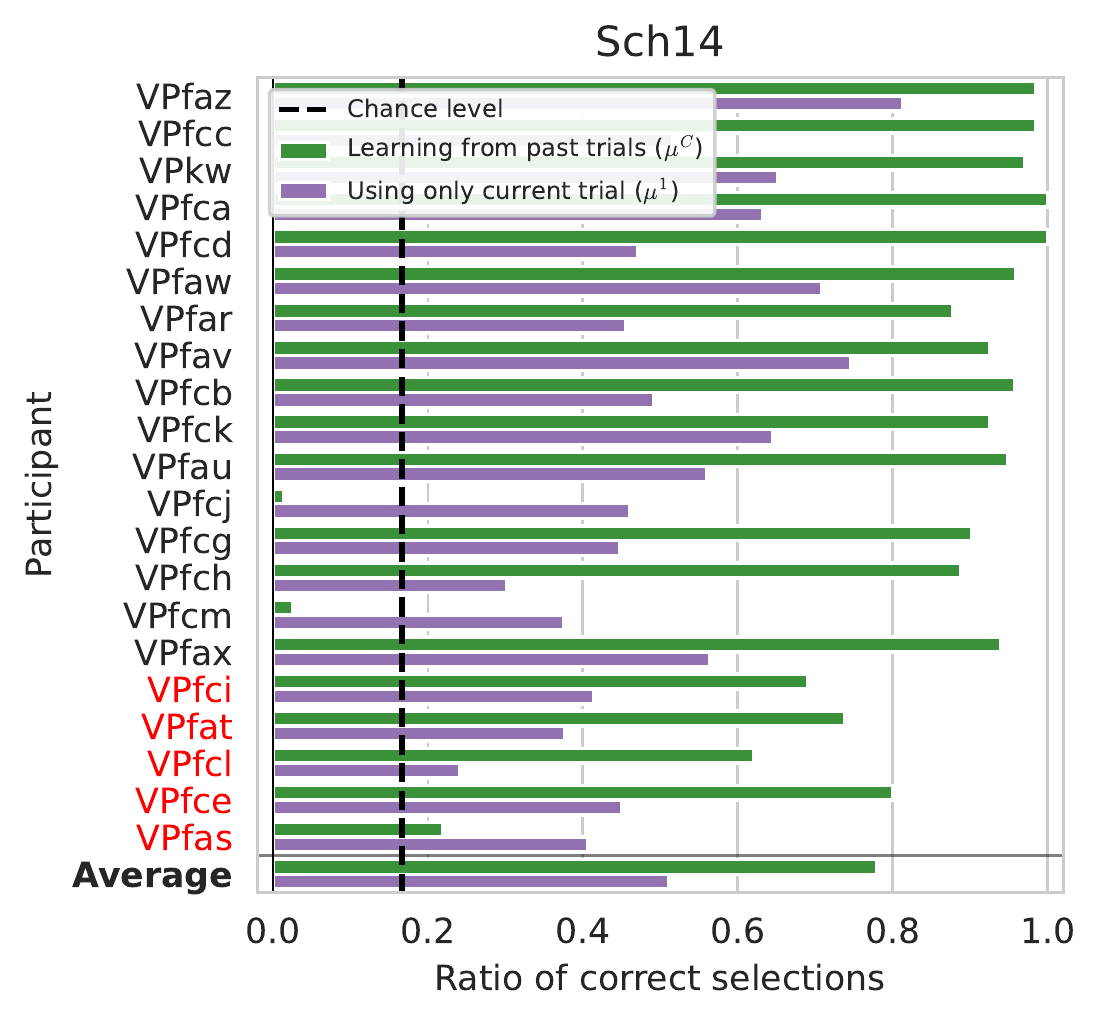}
   \caption{Bar plot with classification rate of the Sch14 dataset. Red-colored participants did not finish the original experiment, as in the original experiment their classification performance was not sufficient to finish spelling the sentence.}
   \label{fig:results_amuse}
\end{figure}

For the auditory dataset in each trial, the attended tone/direction (one out of six) had to be decoded instead of the actual letter, while selecting a letter required at least two trials. As the spelling interface allowed to delete wrong letter and undo wrong first step decisions, the number of trials performed for spelling the same text was different for each participant.
As shown in~\Cref{fig:results_amuse}, the overall performance is worse for this auditory paradigm than for the visual speller datasets.
Nevertheless, UMM worked flawlessly or almost flawlessly for some participants.
Interestingly, the two participants VPfcj and VPfcm perform below chance level with UMM when using information from previous trials.
This is explained by the stimulation protocol, which delivers tone stimuli one after the other instead of in parallel:
When UMM chooses the wrong target assignment in the first trials, the mean estimation in subsequent trials will move the class means along the opposite direction (cf.~\Cref{sec:datasets}).
In the visual speller setups, where one stimulation event highlights multiple symbols, this problem is alleviated, especially when the set of highlighted letters is chosen (pseudo-)randomly (datasets: Hüb17, Hüb18, Lee19) and is not constrained to rows or columns (dataset: Ric13).

\subsection{Improvement over State of the Art Unsupervised Classifiers}

To relate UMM with state of the art decoding methods in BCI, we compare our results with the original publications of the Hüb17 and Hüb18 datasets, as in these datasets the used online classifier was indeed unsupervised.
We only report for the setting where the covariance is estimated using a Toeplitz structure ($\Sigma_t$) and the mean was estimated based on previous confidence values ($\mu^C$).

\begin{figure}[t]
   \centering
   \includegraphics[width=\columnwidth]{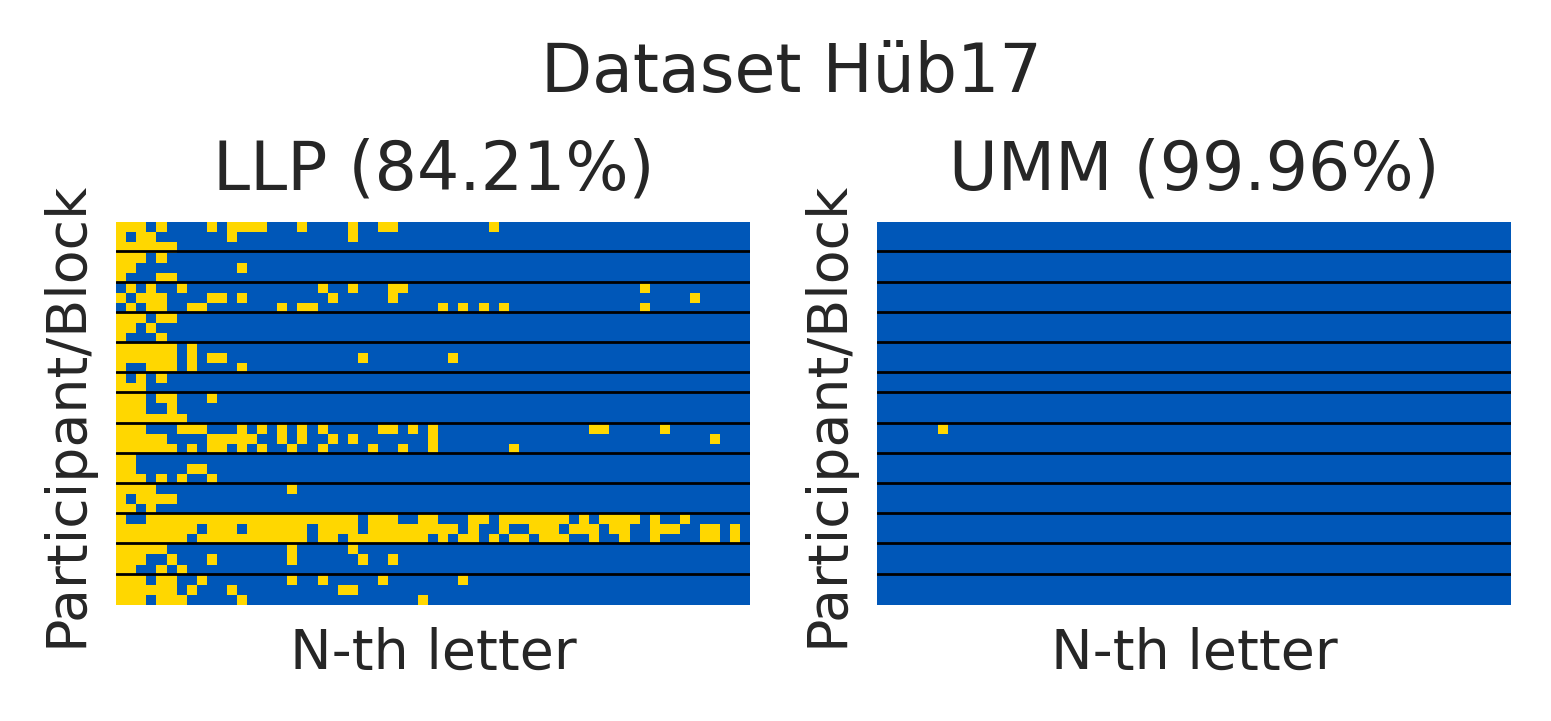}
   \caption{
Heatmap of correctly (blue) and incorrectly (yellow) classified letters for the original learning from label proportions (LLP) method (left) and the proposed novel UMM  method (right). Each row is one block of 63 letters of one participant. Black lines delineate the 13 participants, who executed different numbers of letter blocks each.}
   \label{fig:umm_vs_llp}
\end{figure}

In~\Cref{fig:umm_vs_llp}, the online classification results obtained by the original learning from label proportions (`LLP') method on the Hüb17 dataset~\citep{hubner2017learning} are compared to the classification results we obtained using UMM.
In this plot, each row corresponds to one block of 63 letters spelled by one participant.
Whereas in the original experiment, a clear ramp-up phase of `LLP' can be observed for all participants, UMM works from the very start for basically every participant.
Even for participant 11 (third last) who had difficulties using the BCI  with `LLP', UMM would allow for a perfect classification performance.
Note that the EEG data of the second block of participant 6 could not be loaded due to missing optical markers and was therefore omitted for both methods.

\begin{figure}[t]
   \centering
   \includegraphics[width=\columnwidth]{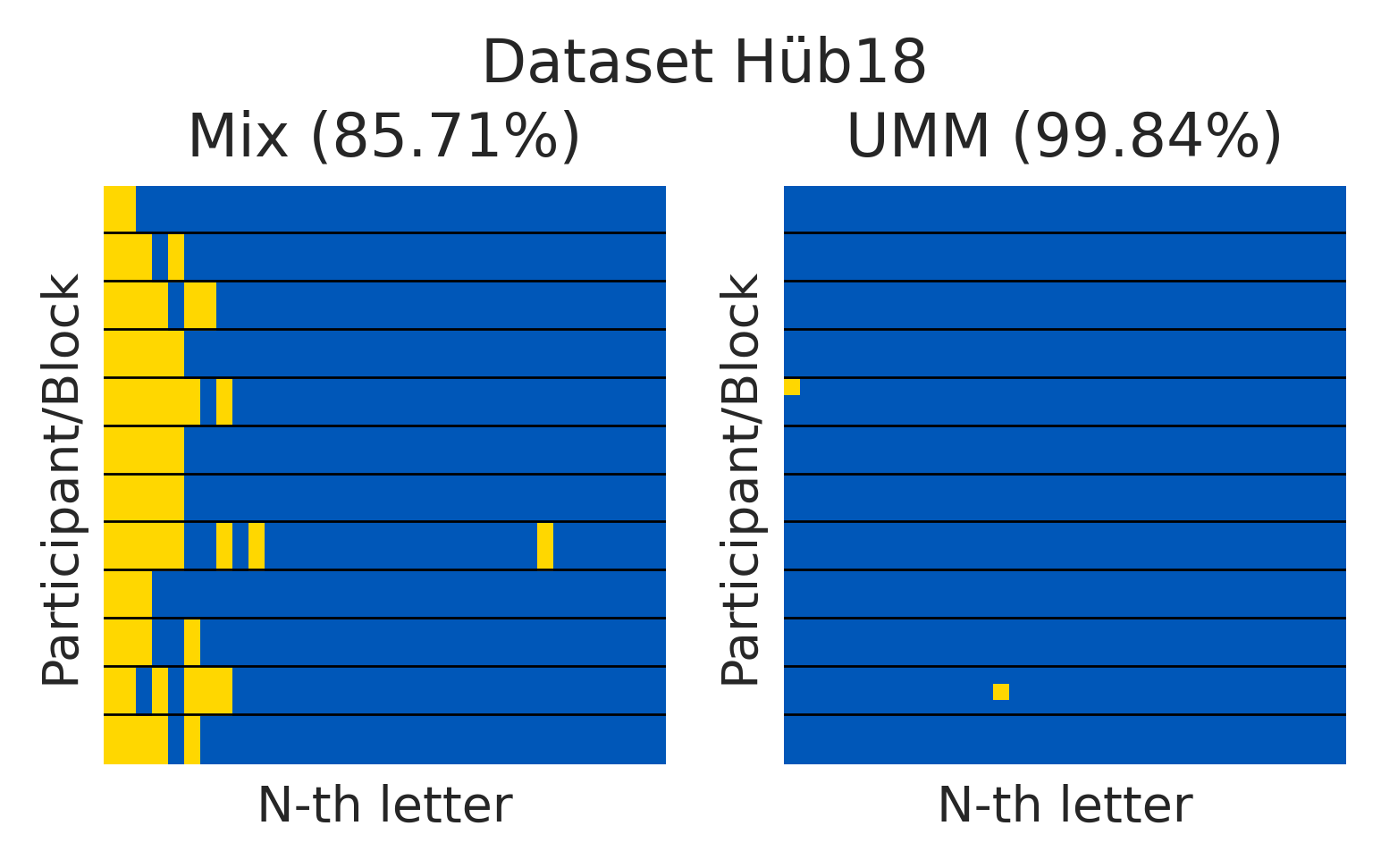}
   \caption{Heatmap of correctly (blue) and incorrectly (yellow) classified letters for the original `Mix' method (left) and UMM (right). Each row is one block of 35 letters of one participant. Black lines delineate the 12 participants, who executed different numbers of letter blocks each. Overall classification performance is given in brackets.}
   \label{fig:umm_vs_mix}
\end{figure}

In the original recording of the Hüb18 dataset, \citet{hubner2018unsupervised} compared learning from label proportions, expectation maximization and their proposed combination thereof, i.e., the so-called `Mix' method.
Note that in this online experiment each block of 35 letters had been classified by one of the three methods.
As we could not match which method had been used for which block, we only report the performance of the `Mix' method, as this method had been reported as the best of the compared approaches by Hübner and colleagues and can still be considered as the state of the art unsupervised classification method for ERP-based BCI.
The results of a comparison between the `Mix' method and UMM are shown in~\Cref{fig:umm_vs_mix}.
In this dataset, no clearly bad performing participants can be observed, but UMM notably again works from the very first letter.
Note that the performance of `Mix' (85.71\%) seems to be only slightly better than `LLP' (84.21\%), but this is caused by the shorter sentences that had to be spelled (unsupervised methods tend to perform badly especially for the first few letters).

Note that in both, the Hüb17 and the Hüb18 datasets, UMM's performance when using no past information at all (see columns $\Sigma^1_t,\mu^1$ and $\Sigma^1_s,\mu^1$ in~\Cref{fig:heatmap_all}), is already very high, especially when the covariance estimation makes use of the block-Toeplitz regularization.

\subsection{Comparison to Supervised Classification}

In the Lee19 dataset, the original authors used the first 33 letters recorded in copy-spell mode as calibration data, while the remaining 36 were actually classified online.
Note that the latter was a copy-spell task as well, however, the interface did not indicate which letter to spell next after every letter.
For this online block, \citet{lee2019eeg} report a mean classification accuracy of $96.8\,\%$.
As UMM is an unsupervised approach, it can not only be applied to the online block, but it would allow online classification already for the first 33 letters. On the full set of 69 letters, UMM reaches a performance of $99.47\,\%$. Evaluating UMM's classification rate only for letters from the online block (letters 34--69, comparable to what Lee and colleagues reported), UMM obtains an accuracy of $99.61\,\%$.

For the Ric13 dataset, the first 15 letters had been used by the original authors for calibration and the remaining 20 had been used for online spelling.
On this dataset, \citet{riccio2013attention} report an average classification accuracy of $97.5\,\%$ which is much better than the $77.86\,\%$ UMM achieves.
However, the bad average classification performance is mainly caused by participants where initial mistakes prohibit UMM from working even when using confidence-base mean estimation.
Detecting and mitigating early mistakes may be key to improve UMM's performance.

\subsection{Reliability of the Confidence Measure}

\begin{figure}[t]
   \centering
   \includegraphics[width=\columnwidth]{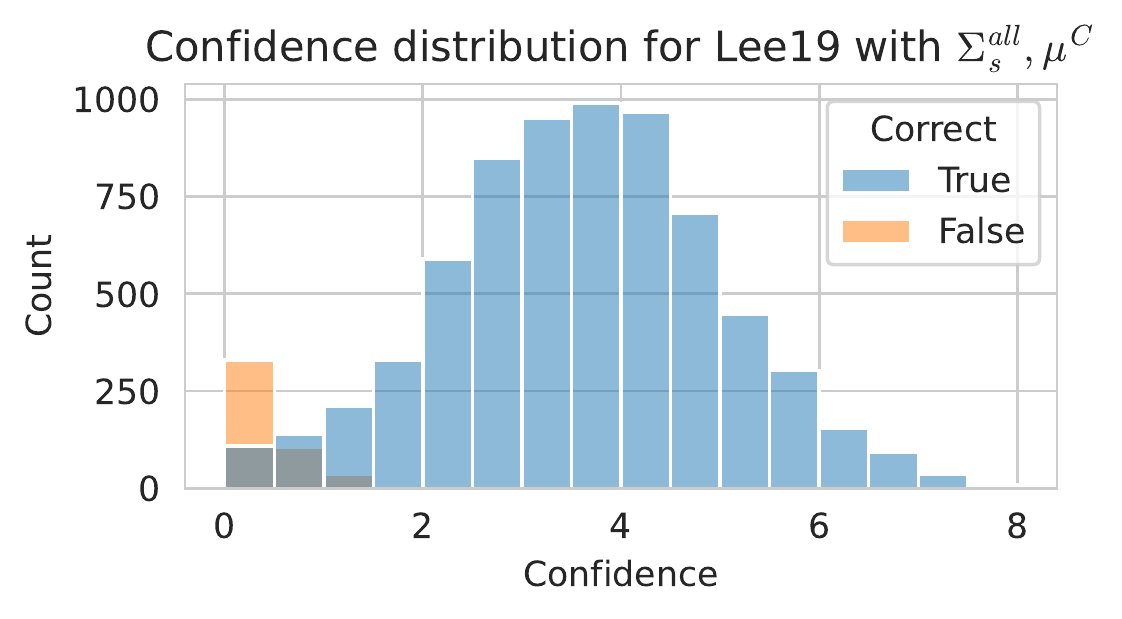}
   \caption{Distribution of confidence values for the Lee19 dataset using confidence-based mean estimation and shrinkage covariance estimation.}
   \label{fig:distribution_lee}
\end{figure}

To assess the reliability of the confidence measure proposed for the unsupervised UMM method, we focus on a setting for which a substantial amount of letters were classified incorrectly.
For this purpose, we chose the traditional shrinkage covariance estimation with confidence-based mean ($\Sigma_s^{all}$, $\mu^C$ classification rate: $93.40\,\%$) of the large Lee19 dataset.
In this setting, the distribution of the confidence values is shown in~\Cref{fig:distribution_lee}.
As expected, the confidence is consistently low for incorrect classifications.
In fact, all wrong classifications had a confidence of $1.5$ or lower, while overall confidence values ranged between $0$ and $7.5$.

\subsection{Confidence as a Predictor of Degenerate Cases}

As mentioned previously, when using past information to estimate the means of a trial, in rare cases it can happen that UMM learns the inverted class means at the start.
This undesired behavior was mainly observed when using the worse shrinkage regularized covariance matrix ($\Sigma_s$).
However, even when using the block-Toeplitz covariance ($\Sigma_t$), we observed the undesired behavior for participants 1,2,4 of the Ric13 dataset and participants VPfcj and VPfcm of the Sch14 dataset.
If this happens, UMM will misclassify reliably below chance level.

\begin{figure}[t]
   \centering
   \includegraphics[width=\columnwidth]{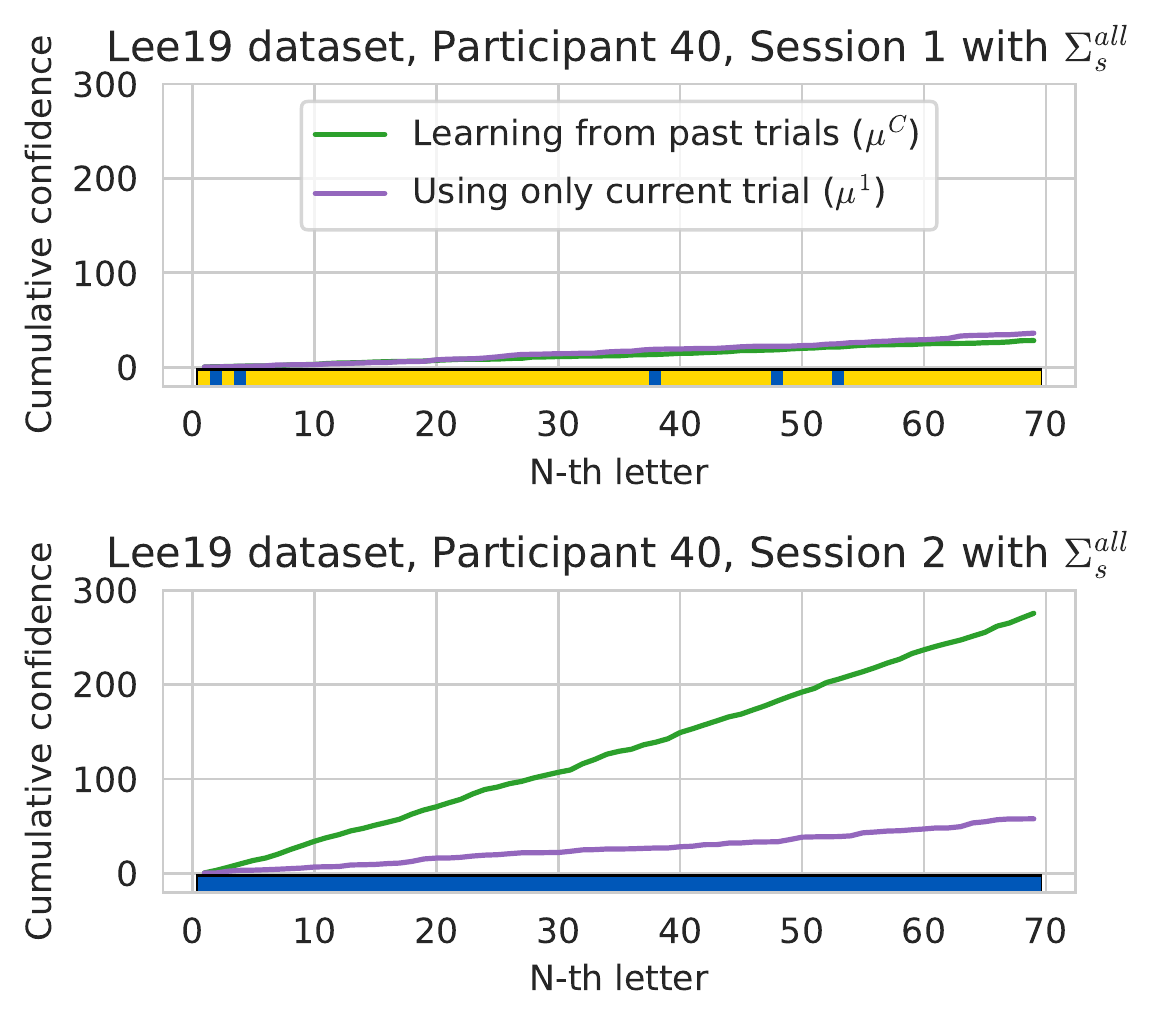}
   \caption{Cumulative confidences for participant 40 of the Lee19 dataset. Blue blocks at the bottom line indicate that the letter was classified correctly using the cumulative mean estimate $\mu^C$, yellow blocks indicate misclassifications.}
   \label{fig:confs_lee}
\end{figure}

We investigated this undesired behavior for the example of participant 40 of the Lee19 dataset in~\Cref{fig:confs_lee} when using shrinkage covariance ($\Sigma_s^{all}$) estimation.
This figure provides the cumulative confidence values, i.e., the sum of all confidences obtained over all trials.
In session 1 (top plot), using UMM with past information would lead to almost no correct classification at all for this participant, whereas in session 2 UMM performs perfectly.
The cumulative confidence can be used to detect this degenerate case in session 1, as the confidence of UMM using $\mu^C$ (green line) is barely different from the confidence that uses $\mu^1$ (purple line).
Contrary, in session 2 $\mu^C$ reliably accumulates to higher confidences.
Note that in the UMM implementation, calculating both confidences corresponds to one additional matrix multiplication only, which has only a negligible run time impact.

\subsection{Using UMM Confidence to Assess the Quality of the Stimulation Sequence}
\label{sec:conf_stim}

In the Hüb17 and Hüb18 datasets, the stimulation sequences were pre-generated before the start of the experiments.
The same sequence was used in every block/sentence of each participant across both datasets.
This means that, for example, when letter `A' was the symbol to be attended in the third position of a sentence, each participant was presented with the same order of highlighting events for this letter in this position.

\begin{figure*}[t]
   \centering
   \includegraphics[width=\textwidth]{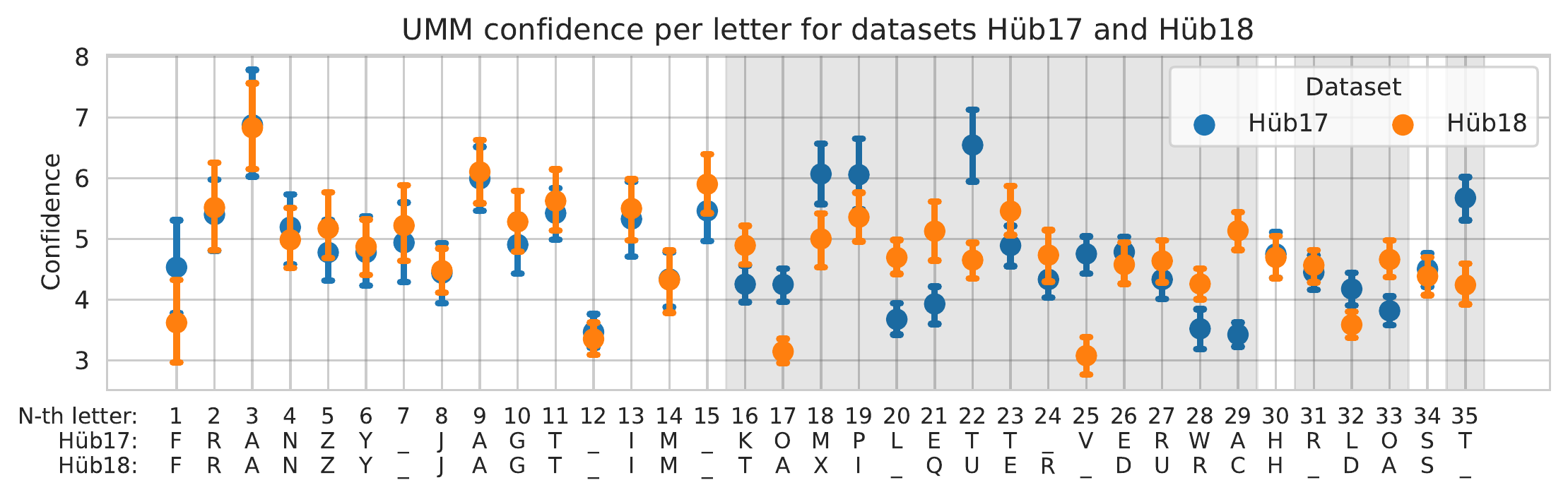}
   \caption{Classifier confidence of the first 35 letters of the Hüb17 and Hüb18 dataset. The sentence a participant had to spell in each dataset is given on the bottom. Areas with a gray background indicate that different letters had to be spelled at that position. Error bars indicate the 95\,\% confidence interval of the mean.}
   \label{fig:confidences_per_letter}
\end{figure*}

The grand average confidence values for each letter for both the Hüb17 and the Hüb18 datasets are visualized in~\Cref{fig:confidences_per_letter}.
By design, the first 15 letters participants had to spell were the same between the two datasets, while this happened by chance for letters 30 and 34 in the later part.
Interestingly the average classifier confidences are virtually equal when the letters are the same at a certain position in the sentence to be spelled, even though the 13 participants in the Hüb17 study were different from the 12 participants from the Mix study.
However, if different letters had to be spelled---i.e., the stimulation sequence of the attended letter is different between Hüb17 and Hüb18 participants---the confidences are not overlapping.
The same holds for the same letter being spelled in different positions of the sentence as then different stimulation sequences have been used.
See for example the confidence values for the space/underscore character in positions 12 and 15, where the average confidence is around $3.4$ and around $5.7$ respectively.
This observation strongly indicates that the confidence of UMM can be used---when other experimental parameters are identical---to identify stimulation sequences which are hard or easy to classify for UMM.

\section{Discussion}
\label{sec:discussion}

We proposed the UMM approach, this unsupervised method would allow healthy participants to use a ERP-based visual speller BCI without any calibration and with almost no misclassifications from the very first letter on.
The experiments on a large number of datasets indicate, that UMM performs better than traditionally used supervised and unsupervised LDA binary classification.
Note that the latter use aggregated classifier outputs to make predictions in a multi-class setting, whereas UMM solves this multi-class directly.
A possible explanation of the observed performance difference is that our method does not try to correctly estimate the parameters of an actual projection direction (while for LDA, the normal vector $\mathbf{w}$ needs to be estimated)---instead, it selects the classification assignment that produces the largest difference between the class means, which is a much simpler task but still sufficient to determine the target ERP stimulus.
Solving this simpler task only may enable UMM to flexibly adapt to changes of the underlying signals, while most decoding approaches cannot cope well with such non-stationarities.
See, e.g., participant 11 (third last participant from the top in~\Cref{fig:umm_vs_llp}), who had a `overall low signal-to-noise ratio (SNR)'~\cite{hubner2017learning} which yields very poor performance using LDA.
If this low SNR is caused by, e.g., latency jitter of the discriminative ERP components between trials, the true ERPs cannot be captured by mere class-wise averaging of the epochs across trials, which is used in traditional classification approaches.
In contrast, UMM will consider the most discriminative dimensions per trial---albeit with more emphasis to the traditional class-wise means when using optimistic or confidence-based mean estimations.

The proposed mean estimation methods of our approach can be interpreted as a regularization of the mean using past trials and na\"ive labeling of these past trials.
Compared to a traditional LDA---here, the data is projected on the direction of the difference between class means without considering the current trial---this allows UMM to consider deviations from the directions of the class means when classifying a new trial.
Note that our current straightforward approaches weight previous trials with the amount of trials available (optimistic mean estimate), which already works well on the majority of participants.
However, as it probably is not optimal, it leaves room for future improvement.

Our proposed UMM framework would also allow a `forgetting of very old trials' for mean and covariance estimation, to cope with non-stationarities.
A rather extreme case is to use UMM instantaneously, e.g., with $\Sigma_t^1$ and ${\mu}^1$.
Here, no past information is used at all, nevertheless this variant achieves up to $96.11\,\%$ classification accuracy (on the Hüb18 dataset) and is completely immune towards all signal non-stationarities between trials (but not within a trial).

In contrast to expectation maximization, which typically is computationally intensive, our proposed approach takes around $0.5$\,s to predict a letter on a i7-8700K CPU (released in 2017) in our provided implementation.
Note that the current implementation has more potential for run time improvements, e.g., by optimizing the inversion of the block-Toeplitz matrix~\cite{poletti2021superfast} which currently makes up more than half of the total time required.
Additionally, \citet{hubner2018unsupervised} observed that expectation maximization sometimes can get stuck in local optima and take many trials before showing a good performance, even on the high quality data of the Mix dataset.
Aside from better classification performance, another benefit of UMM over the `Mix' (which is a combination of `LLP' and expectation maximization) method is that it does not require a paradigm modification: it is readily applicable to any existing ERP-based BCI paradigm we can think of, which requires a multi-class selection by solving multiple binary classification problems, which is almost always true.
However, UMM can benefit from suitable paradigm design, for example, it tends to perform better when the set of highlighted letters is not constrained to form a row or column but is be chosen as (pseudo-)random subsets.

From a neuroscience perspective, it is also interesting to investigate which stimuli or epochs, are easier or harder to classify for UMM in order to assess the elicited ERPs.
However, UMM can only be used to predict the final one-out-of-multiple-classes letter (or another multi-class label) directly and cannot directly operate on a single stimulus.
Still, as UMM calculates a mean estimate and the inverted covariance matrix, it is trivial to simply use these to obtain an LDA ($\vec{w}=\Sigma^{\shortminus 1} \Delta \vec{\mu}$), which could then be used for actual binary classification of individual epochs.

While for more difficult data (Ric13 dataset, with patients and a row / column layout) or generally for weaker ERP responses (Sch14 dataset with auditory evoked ERPs) it rarely can happen that UMM fails to perform at all when using mean information from past trials.
This is a known phenomenon in setups using na\"ive labeling~\cite{kuncheva2008case-study} when early trials are misclassified.
This emphasizes the need for methods that work even when extremely limited data is available---for example, using the block-Toeplitz structured covariance matrix---to reliably classify the first few trials/letters in a BCI experiment.
Using the proposed cumulative confidence measure (see~\Cref{fig:confs_lee}), these degenerate cases could be detected in an online experiment, such that UMM can be informed to, e.g., discard the mean information obtained so far and start over.
A different approach to cope with harder datasets could be to initialize the UMM mean estimation either with prior knowledge, or using a short calibration phase, however, this would make UMM a supervised method.
As the UMM method appears to perform always above chance level when no past information is used, for example as shown in~\Cref{fig:results_amuse}, this information could be used to obtain a robust mean estimate overall and may also serve to prevent the undesired cases where UMM performs below chance level.
Finally, UMM could also make use of a recalculation procedure similar to the post-hoc re-analysis proposed by~\citet{hubner2017learning}.
The authors make use of recent information (i.e., better mean and covariance estimates the more data is recorded) to re-classify previous trials.
If this rectifies mistakes on the first few trials, the corrected aggregated mean estimate (confidence-based or optimistic) will become more reliable in future trials.

\section{Conclusion}
\label{sec:conclusion}

We introduced the simple Unsupervised Mean-difference Maximization (UMM) method for ERP-based BCI systems.
It does not require labels, i.e., no calibration phase is needed.
UMM delivers a highly competitive classification accuracy over multiple visual speller datasets with healthy participants ($99.96\,\%$, $99.84\,\%$ and $99.47\%$).
Rare shortcomings were observed in a patient dataset ($77.86\,\%$) and an auditory dataset ($82.52\,\%$), which can be detected using a proposed confidence metric, which comes basically for free when using UMM.
For BCI practitioners it is important to emphasize that UMM can be applied to any available ERP-based BCI protocol, but benefits from suitable paradigm design.
Practitioners should consider incorporating UMM into their BCI systems to eliminate the need for calibration as well as to allow participants to instantly be able to spell.

\section*{Disclaimer}

There is currently a patent application pending for applications using UMM.

\section*{Acknowledgements}

Our work was supported by the German Research Foundation project SuitAble (DFG, grant number 387670982) and by the Federal Ministry of Education and Research (BMBF, grant number 16SV8012). The authors would also like to acknowledge support by the state of Baden-W\"{u}rttemberg, Germany, through bwHPC and the German Research Foundation (DFG, INST 39/963-1 FUGG).

\bibliography{bib/master}

\begin{thebibliography}{31}
\providecommand{\natexlab}[1]{#1}
\providecommand{\url}[1]{\texttt{#1}}
\expandafter\ifx\csname urlstyle\endcsname\relax
  \providecommand{\doi}[1]{doi: #1}\else
  \providecommand{\doi}{doi: \begingroup \urlstyle{rm}\Url}\fi

\bibitem[Barachant \& Congedo(2014)Barachant and
  Congedo]{barachant2014plugplay}
Barachant, A. and Congedo, M.
\newblock A plug\&play {P300} {BCI} using information geometry.
\newblock \emph{arXiv preprint arXiv:1409.0107}, 2014.

\bibitem[Blankertz et~al.(2011)Blankertz, Lemm, Treder, Haufe, and
  M{\"u}ller]{blankertz2011single-trial}
Blankertz, B., Lemm, S., Treder, M., Haufe, S., and M{\"u}ller, K.-R.
\newblock Single-trial analysis and classification of {ERP} components---a
  tutorial.
\newblock \emph{NeuroImage}, 56\penalty0 (2):\penalty0 814--825, 2011.

\bibitem[Cecotti \& Graser(2010)Cecotti and Graser]{cecotti2010convolutional}
Cecotti, H. and Graser, A.
\newblock Convolutional neural networks for {P300} detection with application
  to brain-computer interfaces.
\newblock \emph{IEEE transactions on pattern analysis and machine
  intelligence}, 33\penalty0 (3):\penalty0 433--445, 2010.

\bibitem[Cohen \& Sances(1977)Cohen and Sances]{cohen1977stationarity}
Cohen, B.~A. and Sances, A.
\newblock Stationarity of the human electroencephalogram.
\newblock \emph{Medical and Biological Engineering and Computing}, 15\penalty0
  (5):\penalty0 513--518, 1977.

\bibitem[Farwell \& Donchin(1988)Farwell and Donchin]{farwell1988talking}
Farwell, L.~A. and Donchin, E.
\newblock Talking off the top of your head: toward a mental prosthesis
  utilizing event-related brain potentials.
\newblock \emph{Electroencephalography and Clinical Neurophysiology},
  70\penalty0 (6):\penalty0 510--523, 1988.

\bibitem[Haider \& Fazel-Rezai(2017)Haider and
  Fazel-Rezai]{haider2017application}
Haider, A. and Fazel-Rezai, R.
\newblock Application of {P300} event-related potential in brain-computer
  interface.
\newblock \emph{Event-Related Potentials and Evoked Potentials}, 1:\penalty0
  19--36, 2017.

\bibitem[Halder et~al.(2015)Halder, Pinegger, K{\"a}thner, Wriessnegger,
  Faller, Antunes, M{\"u}ller-Putz, and K{\"u}bler]{halder2015brain-controlled}
Halder, S., Pinegger, A., K{\"a}thner, I., Wriessnegger, S.~C., Faller, J.,
  Antunes, J. B.~P., M{\"u}ller-Putz, G.~R., and K{\"u}bler, A.
\newblock Brain-controlled applications using dynamic {P300} speller matrices.
\newblock \emph{Artificial intelligence in medicine}, 63\penalty0 (1):\penalty0
  7--17, 2015.

\bibitem[H{\"u}bner(2020)]{hubner2020from}
H{\"u}bner, D.
\newblock \emph{From Supervised to Unsupervised Machine Learning Methods for
  Brain-Computer Interfaces and Their Application in Language Rehabilitation}.
\newblock PhD thesis, University of Freiburg, 2020.

\bibitem[H{\"u}bner et~al.(2017)H{\"u}bner, Verhoeven, Schmid, M{\"u}ller,
  Tangermann, and Kindermans]{hubner2017learning}
H{\"u}bner, D., Verhoeven, T., Schmid, K., M{\"u}ller, K.-R., Tangermann, M.,
  and Kindermans, P.-J.
\newblock Learning from label proportions in brain-computer interfaces: online
  unsupervised learning with guarantees.
\newblock \emph{PloS one}, 12\penalty0 (4), 2017.

\bibitem[H{\"u}bner et~al.(2018)H{\"u}bner, Verhoeven, M{\"u}ller, Kindermans,
  and Tangermann]{hubner2018unsupervised}
H{\"u}bner, D., Verhoeven, T., M{\"u}ller, K.-R., Kindermans, P.-J., and
  Tangermann, M.
\newblock Unsupervised learning for brain-computer interfaces based on
  event-related potentials: Review and online comparison.
\newblock \emph{IEEE Computational Intelligence Magazine}, 13\penalty0
  (2):\penalty0 66--77, 2018.

\bibitem[Jayaram et~al.(2016)Jayaram, Alamgir, Altun, Scholkopf, and
  Grosse-Wentrup]{jayaram2016transfer}
Jayaram, V., Alamgir, M., Altun, Y., Scholkopf, B., and Grosse-Wentrup, M.
\newblock Transfer learning in brain-computer interfaces.
\newblock \emph{IEEE Computational Intelligence Magazine}, 11\penalty0
  (1):\penalty0 20--31, 2016.

\bibitem[Jung et~al.(1998)Jung, Makeig, Westerfield, Townsend, Courchesne, and
  Sejnowski]{jung1998analyzing}
Jung, T.-P., Makeig, S., Westerfield, M., Townsend, J., Courchesne, E., and
  Sejnowski, T.~J.
\newblock Analyzing and visualizing single-trial event-related potentials.
\newblock \emph{Advances in neural information processing systems}, 11, 1998.

\bibitem[Kaufmann et~al.(2011)Kaufmann, Schulz, Gr{\"u}nzinger, and
  K{\"u}bler]{kaufmann2011flashing}
Kaufmann, T., Schulz, S., Gr{\"u}nzinger, C., and K{\"u}bler, A.
\newblock Flashing characters with famous faces improves {ERP}-based
  brain--computer interface performance.
\newblock \emph{Journal of neural engineering}, 8\penalty0 (5):\penalty0
  056016, 2011.

\bibitem[Kindermans et~al.(2012)Kindermans, Verstraeten, and
  Schrauwen]{kindermans2012bayesian}
Kindermans, P.-J., Verstraeten, D., and Schrauwen, B.
\newblock A {Bayesian} model for exploiting application constraints to enable
  unsupervised training of a {P300}-based {BCI}.
\newblock \emph{PloS one}, 7\penalty0 (4):\penalty0 e33758, 2012.

\bibitem[Kuncheva et~al.(2008)Kuncheva, Whitaker, and
  Narasimhamurthy]{kuncheva2008case-study}
Kuncheva, L.~I., Whitaker, C.~J., and Narasimhamurthy, A.
\newblock A case-study on naive labelling for the nearest mean and the linear
  discriminant classifiers.
\newblock \emph{Pattern Recognition}, 41\penalty0 (10):\penalty0 3010--3020,
  2008.

\bibitem[Ledoit \& Wolf(2004)Ledoit and Wolf]{ledoit2004well-conditioned}
Ledoit, O. and Wolf, M.
\newblock A well-conditioned estimator for large-dimensional covariance
  matrices.
\newblock \emph{Journal of multivariate analysis}, 88\penalty0 (2):\penalty0
  365--411, 2004.

\bibitem[Lee et~al.(2019)Lee, Kwon, Kim, Kim, Lee, Williamson, Fazli, and
  Lee]{lee2019eeg}
Lee, M.-H., Kwon, O.-Y., Kim, Y.-J., Kim, H.-K., Lee, Y.-E., Williamson, J.,
  Fazli, S., and Lee, S.-W.
\newblock {EEG} dataset and {OpenBMI} toolbox for three {BCI} paradigms: An
  investigation into {BCI} illiteracy.
\newblock \emph{GigaScience}, 8\penalty0 (5):\penalty0 giz002, 2019.

\bibitem[Li et~al.(2013)Li, Pan, Wang, and Yu]{li2013hybrid}
Li, Y., Pan, J., Wang, F., and Yu, Z.
\newblock A hybrid {BCI} system combining {P300} and {SSVEP} and its
  application to wheelchair control.
\newblock \emph{IEEE Transactions on Biomedical Engineering}, 60\penalty0
  (11):\penalty0 3156--3166, 2013.

\bibitem[Lin et~al.(2018)Lin, Zhang, Zeng, Tong, and Yan]{lin2018novel}
Lin, Z., Zhang, C., Zeng, Y., Tong, L., and Yan, B.
\newblock A novel {P300} {BCI} speller based on the triple {RSVP} paradigm.
\newblock \emph{Scientific reports}, 8\penalty0 (1):\penalty0 3350, 2018.

\bibitem[Lotte et~al.(2018)Lotte, Bougrain, Cichocki, Clerc, Congedo,
  Rakotomamonjy, and Yger]{lotte2018review}
Lotte, F., Bougrain, L., Cichocki, A., Clerc, M., Congedo, M., Rakotomamonjy,
  A., and Yger, F.
\newblock A review of classification algorithms for {EEG}-based brain--computer
  interfaces: a 10 year update.
\newblock \emph{Journal of neural engineering}, 15\penalty0 (3):\penalty0
  031005, 2018.

\bibitem[Lumley et~al.(2002)Lumley, Diehr, Emerson, Chen,
  et~al.]{lumley2002importance}
Lumley, T., Diehr, P., Emerson, S., Chen, L., et~al.
\newblock The importance of the normality assumption in large public health
  data sets.
\newblock \emph{Annual review of public health}, 23\penalty0 (1):\penalty0
  151--169, 2002.

\bibitem[Mane et~al.(2020)Mane, Chouhan, and Guan]{mane2020bci}
Mane, R., Chouhan, T., and Guan, C.
\newblock {BCI} for stroke rehabilitation: motor and beyond.
\newblock \emph{Journal of Neural Engineering}, 17\penalty0 (4):\penalty0
  041001, 2020.

\bibitem[Musso et~al.(2022)Musso, H\"{u}bner, Schwarzkopf, Bernodusson, LeVan,
  Weiller, and Tangermann]{musso2022aphasia}
Musso, M., H\"{u}bner, D., Schwarzkopf, S., Bernodusson, M., LeVan, P.,
  Weiller, C., and Tangermann, M.
\newblock Aphasia recovery by language training using a brain-computer
  interface -- a proof-of-concept study.
\newblock \emph{Brain Communications}, 2022.
\newblock \doi{10.1093/braincomms/fcac008}.

\bibitem[Nunez(2012)]{nunez2012electric}
Nunez, P.~L.
\newblock Electric and magnetic fields produced by the brain.
\newblock \emph{Brain-Computer Interfaces: Principles and Practice}, pp.\
  171--212, 2012.

\bibitem[Poletti \& Teal(2021)Poletti and Teal]{poletti2021superfast}
Poletti, M.~A. and Teal, P.~D.
\newblock A superfast {Toeplitz} matrix inversion method for single-and
  multi-channel inverse filters and its application to room equalization.
\newblock \emph{IEEE/ACM Transactions on Audio, Speech, and Language
  Processing}, 29:\penalty0 3144--3157, 2021.

\bibitem[Riccio et~al.(2013)Riccio, Simione, Schettini, Pizzimenti, Inghilleri,
  Olivetti~Belardinelli, Mattia, and Cincotti]{riccio2013attention}
Riccio, A., Simione, L., Schettini, F., Pizzimenti, A., Inghilleri, M.,
  Olivetti~Belardinelli, M., Mattia, D., and Cincotti, F.
\newblock Attention and {P300}-based {BCI} performance in people with
  amyotrophic lateral sclerosis.
\newblock \emph{Frontiers in human neuroscience}, 7:\penalty0 732, 2013.

\bibitem[Schreuder(2014)]{schreuder2014towards}
Schreuder, E.-J.~M.
\newblock \emph{Towards efficient auditory {BCI} through optimized paradigms
  and methods}.
\newblock Technische Universitaet Berlin (Germany), 2014.

\bibitem[Sosulski \& Tangermann(2022)Sosulski and
  Tangermann]{sosulski2022introducing}
Sosulski, J. and Tangermann, M.
\newblock Introducing block-{Toeplitz} covariance matrices to remaster linear
  discriminant analysis for event-related potential brain-computer interfaces.
\newblock \emph{Journal of Neural Engineering}, 2022.
\newblock \doi{10.1088/1741-2552/ac9c98}.

\bibitem[Verbaarschot et~al.(2021)Verbaarschot, Tump, Lutu, Borhanazad,
  Thielen, van~den Broek, Farquhar, Weikamp, Raaphorst, Groothuis,
  et~al.]{verbaarschot2021visual}
Verbaarschot, C., Tump, D., Lutu, A., Borhanazad, M., Thielen, J., van~den
  Broek, P., Farquhar, J., Weikamp, J., Raaphorst, J., Groothuis, J.~T., et~al.
\newblock A visual brain-computer interface as communication aid for patients
  with amyotrophic lateral sclerosis.
\newblock \emph{Clinical Neurophysiology}, 132\penalty0 (10):\penalty0
  2404--2415, 2021.

\bibitem[Verhoeven et~al.(2017)Verhoeven, H{\"u}bner, Tangermann, M{\"u}ller,
  Dambre, and Kindermans]{verhoeven2017improving}
Verhoeven, T., H{\"u}bner, D., Tangermann, M., M{\"u}ller, K.-R., Dambre, J.,
  and Kindermans, P.-J.
\newblock Improving zero-training brain-computer interfaces by mixing model
  estimators.
\newblock \emph{Journal of neural engineering}, 14\penalty0 (3):\penalty0
  036021, 2017.

\bibitem[Wolpaw \& Wolpaw(2012)Wolpaw and Wolpaw]{wolpaw2012brain-computer}
Wolpaw, J. and Wolpaw, E.~W.
\newblock \emph{Brain-computer interfaces: principles and practice}.
\newblock OUP USA, 2012.

\end{thebibliography}
\bibliographystyle{preprint_bib}

\end{document}